\documentclass[journal]{IEEEtran}
\usepackage{ifpdf}
\usepackage{cite}
\ifCLASSINFOpdf
\usepackage[pdftex]{graphicx}
\else
\usepackage[dvips]{graphicx}
\fi
\usepackage{amsmath}
\usepackage{algorithmic}
\usepackage{array}

\usepackage{threeparttable}

\usepackage{fixltx2e}
\usepackage{dblfloatfix}
\ifCLASSOPTIONcaptionsoff
\usepackage[nomarkers]{endfloat}
\let\MYoriglatexcaption\caption
\renewcommand{\caption}[2][\relax]{\MYoriglatexcaption[#2]{#2}}
\fi
\usepackage{url}
\usepackage{algorithm}
\usepackage{amsmath}
\usepackage{graphicx}
\usepackage{subfigure}
\usepackage{diagbox}
\usepackage{multirow}
\usepackage{bm}
\usepackage{amsfonts}
\usepackage{color}
\usepackage{verbatim}
\usepackage{amssymb}
\usepackage{amssymb}
\usepackage{pifont}
\usepackage{booktabs}
\usepackage{bbding}

\begin{document}
	\title{Over-sampling De-occlusion Attention Network for Prohibited Items Detection in Noisy X-ray Images}
	\author{Renshuai Tao, Yanlu Wei, Hainan Li, Aishan Liu, Yifu Ding, Haotong Qin, Xianglong~Liu\textsuperscript{*}
		
		%\thanks{* Equal contribution.}
		
		\thanks{R. Tao, Y. Wei, H. Li, A. Liu, Y. Ding, H. Qin, and X, Liu are with the State Key Lab of Software Development Environment, Beihang University, Beijing 100191, China. H. Qin is also with Shen Yuan Honors College, Beihang University. X. Liu is also with Beijing Advanced Innovation Center for Big Data-Based Precision Medicine ({*} Corresponding author: Xianglong Liu, xlliu@nlsde.buaa.edu.cn)}}
	

	\markboth{IEEE Transactions on Multimedia,~Vol.~, No.~, February~2021}%
	{Shell \MakeLowercase{\textit{Tao et al.}}: Over-sampling De-occlusion Attention Network for Prohibited Items Detection in Noisy X-ray Images}



	\maketitle
	
	\begin{abstract}
		Security inspection is X-ray scanning for personal belongings in suitcases, which is significantly important for the public security but highly time-consuming for human inspectors. Fortunately, deep learning has greatly promoted the development of computer vision, offering a possible way of automatic security inspection. However, items within a luggage are randomly overlapped resulting in noisy X-ray images with heavy occlusions. Thus, traditional CNN-based models trained through common image recognition datasets fail to achieve satisfactory performance in this scenario.% and . As a result, traditional CNN-based models learns directly from noisy data tends to have poor performance.
		%Through this issue is very significant, the performance of traditional CNN-based models are far away from satisfactory due to the occlusion problem caused by that.
		To address these problems, we contribute the first high-quality prohibited X-ray object detection dataset named OPIXray, which contains 8885 X-ray images from 5 categories of the widely-occurred prohibited item ``cutters''. The images are gathered from an airport and these prohibited items are annotated manually by professional inspectors, which can be used as a benchmark for model training and further facilitate future research. To better improve occluded X-ray object detection, we further propose an over-sampling de-occlusion attention network (DOAM-O), which consists of a novel de-occlusion attention module and a new over-sampling training strategy. Specifically, our de-occlusion module, namely DOAM, simultaneously leverages the different appearance information of the prohibited items; the over-sampling training strategy forces the model to put more emphasis on these hard samples consisting these items of high occlusion levels, which is more suitable for this scenario. We comprehensively evaluated DOAM-O on the OPIXray dataset, which proves that our model can stably improve the performance of the famous detection models such as SSD, YOLOv3, and FCOS, and outperform many extensively-used attention mechanisms. ~\footnote{Our code can be found at \url{https://github.com/OPIXray-author/OPIXray}.}
	\end{abstract}
	

	% \fi
	\IEEEpeerreviewmaketitle

	\section{Introduction}
	\IEEEPARstart{W}{ith} the increasing crowd density in public transportation hubs, security inspection is widely adopted to examine the passengers' belongings with X-ray scanners, making sure that no prohibited item is mixed into the luggage.
	Usually, items in the luggage are intensely occluded since they are randomly stacked and heavily overlapped with each other.
	Therefore, the prolonged time of inspecting in a myriad of complex X-ray images with unremitting attention may tire the security inspectors to accurately distinguish all the prohibited items, which may lead to public danger.
	Therefore, it is indispensible to adopt an efficient, precise, and automatic approach to assist inspectors in checking the passengers' belongings in X-ray images. 
	%Also, it's not advisable to change shifts frequently for the huge cost of human resources. 
	\begin{figure}[!t]
		\centering
		\includegraphics[width=\linewidth]{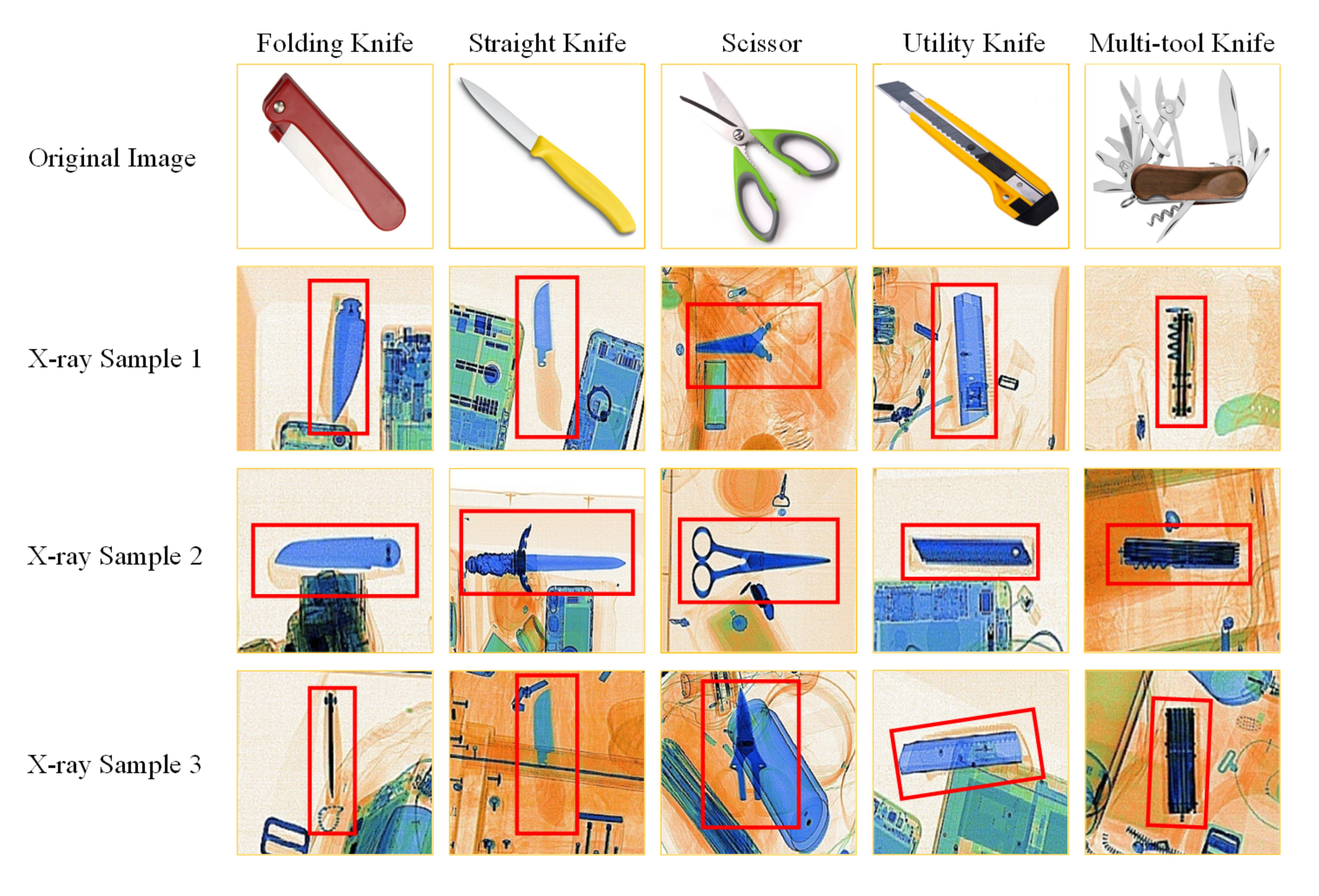}
		\caption{Samples of the five categories of cutters in OPIXray dataset and their corresponding noisy X-ray images.}
		\label{Samples_five}
	\end{figure}

	\begin{table*}[!t]
	\begin{center}
		\newcommand{\tabincell}[2]{\begin{tabular}{@{}#1@{}}#2\end{tabular}}
		\setlength{\tabcolsep}{1.5mm}
		\caption{Comparison of existing open-source X-ray datasets.}
		{
			\small
			\begin{tabular}{lccccccccc}
				\toprule
				\multirow{2}{*}{Dataset} & \multirow{2}{*}{Year} & \multirow{2}{*}{Color} & \multirow{2}{*}{Category} & \multirow{2}{*}{Positive Samples} & \multicolumn{3}{c}{Annotation}  & \multirow{2}{*}{Task} & \multirow{2}{*}{Data Source}\\
				\cmidrule{6-8}
				& && && Bounding box & Number &
				Professional & &\\
				\midrule
				GDXray \cite{mery2015gdxray} & 2015& Gray-scale & 3 & 8,150  & \footnotesize\Checkmark & 8,150  & \footnotesize\XSolidBrush  & Detection & Unknown\\
				SIXray \cite{miao2019sixray} & 2019 &RGB& 6 & 8,929 & \footnotesize\XSolidBrush& \footnotesize\XSolidBrush& \footnotesize\XSolidBrush & Classification & Subway Station\\
				\midrule
				\textbf{OPIXray} & 2020&RGB & 5 & 8,885 & \footnotesize\Checkmark & 8,885   & \footnotesize\Checkmark  & Detection & Airport\\
				\bottomrule
			\end{tabular}
		}
	\end{center}
	\label{dataset-comparison}
\end{table*}
	Fortunately, the innovation of deep learning, especially the convolutional neural network, makes it possible to accomplish the goal by transferring it into object detection task in computer vision \cite{sultana2020unsupervised,wu2020deeper,li2020stacked,ren2020salient,jiang2020cmsalgan}. However, different from traditional detection tasks, in this scenerio, items within a luggage are randomly overlapped where most areas of objects are occluded resulting in noisy X-ray images with heavy occlusions. Thus, this characteristic of these noisy X-ray images leads to strong requirement of high-quality datasets and models with satisfactory performance for this detection task. 
	
	Regarding dataset, to the best of our knowledge, there are two released X-ray benchmarks for classification task, namely GDXray \cite{mery2015gdxray} and SIXray \cite{miao2019sixray}, which are incommensurate with our target of occluded prohibited item detection. Additionally, all images in GDXray are gray-scale, and less than 1\% pictures in SIXray are annotated with prohibited items, both of which are not of very high quality. In addition, traditional CNN-based models \cite{li2021image,tang2019salient,qiu2020hierarchical,xiao2018deep,fu2018refinet,qin2020forward} trained through common image recognition datasets fail to achieve satisfactory performance in this scenario. This urgently requires researchers to make breakthroughs in both datasets and models.
	
	%As for the models, traditional CNN-based detection models fail to achieve satisfactory performance in X-ray prohibited items detection due to heavy occlusions. In particular, the models aiming to solve the occlusion problem in other areas, such as person re-identification \cite{zhang2018occlusion,zhou2018bi,wang2018repulsion} and face recognition \cite{ge2017detecting,yang2016nuclear,song2019occlusion}, are not applicable for this scenario due to the particularity of X-ray images. Object occlusion in X-ray images for security inspection often exists between prohibited items and other safety items, which pertains to inter-class occlusion while these senses belongs to intra-class occlusion.

	%In our prior conference publication \cite{WeiOccluded2020},
	In this work, we contributed the first high-quality dataset named OPIXray, which aims at occluded prohibited items detection in security inspection. Considering that cutter is the most common tool passengers carry, we choose it as the prohibited item to detect. As a result, OPIXray contains 8885 X-ray images of 5 categories of cutters (illustrated in Fig. \ref{Samples_five}). Besides, we provide abundant analysis and discussions of the OPIXray dataset including the construction principles, the quality control procedures, and the potential tasks. In order to simulate the real security inspection scene to the greatest extent, the images of the dataset are gathered from an airport and these prohibited items are annotated manually by professional inspectors, which can be used as a benchmark for model training and further facilitate future research.
	
	%The quality of the dataset is an essential problem because more rubust and sophisticated models can hardly be derived from coarse dataset. By exploiting images in high quality dataset, better models and algorithms can be vigorously provoked, reasonably evaluated, and fairly compared.
	In addition, to further improve occluded X-ray object detection, we propose an over-sampling de-occlusion attention network (DOAM-O), which consists of a de-occlusion attention module (DOAM) and an new over-sampling training strategy. DOAM is a plug-and-play module, which can be easily inserted to promote most popular detectors. DOAM simultaneously leverages the different appearance information (shape and material) of the prohibited item to generate the attention map, which helps refine feature maps for the general detectors. The over-sampling training strategy retrains these hard samples (i.e., prohibited items containing high-level occlusions) to extract more effective information from the hard X-ray samples, further boosting the performance of the model. To the best of our knowledge, this is the first work that explicitly exploits over-sample technologies to the X-ray prohibited items, which might offer a new way to solve the occlusion detection problem under noisy data scenarios.

	Note that in this paper, we extend our prior conference publication \cite{WeiOccluded2020} which contributes the rudiment of the OPIXray dataset and proposes a de-occlusion network which mainly consists of the DOAM module with traditional training strategy (DOAM-T). Compared with our previous conference paper, we add more analysis and discussions of the OPIXray dataset including the construction principles, the potential tasks, etc., to make the process of constructing this dataset clearer. We hope these investigations can be beneficial to facilitate the expansion and scale-up of OPIXray in the future research as well as the development of prohibited items detection in X-ray security inspection. Besides, we extend the model of our previous conference paper by replacing the traditional training strategy of the model with a new over-sampling training strategy (DOAM-O). The new strategy could make the model learn to put more emphasis on these hard samples the existed massively under the X-ray detection dataset, which is more suitable for the prohibited items detection scenario. Experiments show that DOAM-O could beat the DOAM-T by a significant margin in most cases, validating our claim that the over-sampling training strategy could help the model put more emphasis on these hard samples. Moreover, we discuss different implementations for the over-sampling training strategy and different attention mechanisms, ensuring our method is theoretically reasonable and yields competitive performance.
	%by simultaneously leveraging the different appearance information of the prohibited items to generate the attention map with traditional training strategy. In this paper, with the newly proposed over-sampling training strategy, 

		Our contributions can be listed as follows:
		\begin{itemize}
			\item{We contribute the first high-quality dataset, which named OPIXray, which aims at occluded prohibited items detection in security inspection. Besides, we introduce abundant analysis and discussions for the OPIXray dataset. We hope that contributing this high-quality dataset can promote the development of prohibited items detection in noisy X-ray images.}
			\item{We propose a plug-and-play module, namely DOAM, which simultaneously leverages the different appearance information (shape and material) of items. DOAM helps refine feature maps for the general detectors, promoting the performance of most popular detectors.}
			\item{We propose an over-sampling training strategy, making models learn to put more emphasis on these hard samples existed massively, which is more suitable for this scenario. To the best of our knowledge, this is the first work exploiting the over-sampling technologies to prohibited items detection, which might provide a new way to solve the occlusion problem in noisy X-ray images.}
			\item{Extensive experiments were conducted on the published dataset OPIXray, which is the only dataset for X-ray prohibited items detection. The results demonstrate that our method can drastically boost the detection accuracy and achieve the new SOTA performance for this task.}
		\end{itemize}
	
	The structure of the paper is illustrated as follows: Section \ref{Section:relatedwork} introduces the related works; Section \ref{Section:framework} introduces the OPIXray dataset we contributed; Section \ref{Section:model} describes and discusses the proposed de-occlusion methodology and over-sampling training strategy; Section \ref{Section:exp} demonstrates the experiments; and Section \ref{Section:conclusion} summarizes the whole contributions and provides the conclusion.

	% I wish you the best of success.
	
	
	
	\section{Related Work}\label{Section:relatedwork}
	% Subsection text here.
	
	%\IEEEpubidadjcol
	\subsection{X-ray Images and Benchmarks}
	X-ray shows great power in numerous tasks, such as medical imaging analysis \cite{guo2019improved,chaudhary2019diagnosis,lu2019towards} and security inspection \cite{miao2019sixray,huang2019modeling,hassan2020detecting}. Notwithstanding, occlusion between items could disturb the accessibility of the information contained in X-ray images.
	
	Several studies in the literature have endeavored for addressing this challenging issue. However, only a few X-ray datasets have been proposed because of the distinctiveness of security inspection. One released benchmark, named GDXray\cite{mery2015gdxray}, contains 19407 images, partial of which embraces prohibited items of three categories, including gun, shuriken and razor blade. However, images in GDXray are gray-scale with a very simple background, which is far away from complex real-world scenario. SIXray\cite{miao2019sixray}, a recently released benchmark, is a large-scale X-ray dataset consisting of 1059231 X-ray images, which is about 100 times larger than the GDXray dataset\cite{mery2015gdxray}. But the positive samples are deficient (less than 1\%) for imitating a comparable real-world testing environment where prohibited items that inspectors target to only show up with low recurrence. Besides, different from ours, SIXray is a dataset for the task of classification, concentrating on the issue of data imbalance.
	
	\subsection{Attention Mechanism}
	Attention is a method which tends to dispose the most instructive parts of signals among all the accessible computational resources. It has been widely-used in numerous scenes, such as action recognition \cite{li2018unified,li2020spatio,hou2017content}, visual question answering \cite{yu2020reasoning,zhang2019frame,huasong2020self}, and adversarial learning \cite{Liu2019Perceptual,hu2020adversarial,Liu2020Spatiotemporal,li2020attentionfgan}. Attention grasps long-range contextual information which can be generally applied in different tasks, including machine translation \cite{vaswani2017attention}, image captioning \cite{chen2017sca}, scene segmentation \cite{fu2019dual} and object recognition \cite{tang2015rgb}. \cite{wang2018non} utilized self-attention module to explore the effectiveness of non-local operation in space-time dimensions for videos and images. \cite{fu2019dual} proposed a dual attention network (DANet) for scene segmentation by capturing contextual dependence based on the self-attention mechanism. Squeeze-and-Excitation Networks (SENet) \cite{hu2018squeeze}, terming the Squeeze-and-Excitation block (SE), models inter-dependencies between channels, which adaptively re-calibrates channel-wise feature.

	\subsection{Object Detection}
	In computer vision, object detection is one of important tasks, which underpins a few instance-level recognition tasks and many downstream applications \cite{zhou2020salient,qiu2020hierarchical,fu2018refinet}. Most of the CNN-based methods can be further divided into two general approaches: proposal-free detectors and proposal-based detectors.
	Recently one-stage methods have gained much attention over two-stage approaches due to their simpler design and competitive performance. Due to that the security inspection has a high demand for time cost, in this paper, we mainly use single-stage models as a comparison. Here we review some work that is the closest to ours. SSD \cite{liu2016ssd} discretizes the output space of bounding boxes into a set of default boxes over different aspect ratios and scales. YOLO \cite{redmon2016you, redmon2017yolo9000, redmon2018yolov3, bochkovskiy2020yolov4} is the collection of a series of well-known methods which values both real-time and accuracy among single-stage detection algorithms. FCOS \cite{tian2019fcos} proposes a fully convolutional one-stage object detector to solve object detection in a per-pixel prediction fashion, analogue to other dense prediction problems such as semantic segmentation.

	\section{The OPIXray Dataset}\label{Section:framework}
	The quality of the dataset is an essential problem because more rubust and sophisticated models can hardly be derived from coarse dataset. By exploiting images in high quality dataset, better models and algorithms can be vigorously provoked, reasonably evaluated, and fairly compared. Needless to say, a professional dataset with high-quality annotations is much more critical in special practice scenarious.
	In this work, we contribute to the first and high-quality dataset aiming at occluded prohibited items detection in security inspection. Next we introduce the construction principles, data properties and potential tasks of the proposed OPIXray dataset.

	\begin{table}[!t]
	\centering
	\renewcommand\arraystretch{1.3}
	\setlength{\tabcolsep}{7pt}
	\caption{The category distribution of the OPIXray dataset.}
	\label{data_table}
	\small
	\begin{tabular}{ccccccc}
		\hline
		\multirow{2}{*}{OPIXray} & \multicolumn{5}{c}{Categories}                     & \multirow{2}{*}{Total} \\ \cline{2-6}
		& FO & ST & SC & UT & MU &                        \\ \hline
		Training                     & 1589    & 809      & 1494    & 1635    & 1612       & 7109                   \\
		Testing                      & 404     & 235      & 369     & 343     & 430        & 1776                   \\ \hline
		Total                            & 1993    & 1044     & 1863    & 1978    & 2042       & 8885                   \\ \hline
	\end{tabular}
\end{table}

	\subsection{Construction Principles}
		We construct the OPIXray dataset in accordance with the following two principles:
		\begin{itemize}
			\item \textbf{Common Category}. The categories of prohibited items of the images should belong to those frequently seen in daily life, which makes the application more practical.
			\item \textbf{Professional Annotation}. Objects in X-ray images are quite different from those in natural images, whose color-monotonous and luster-insufficient characteristics make them difficult to be recognized for people without professional training. Therefore, the boundary box of prohibited items in X-ray images should be annotated professionally.
			\item \textbf{Quality Control Procedures}. The construction of data sets needs to include the necessary quality control procedures, making further data collection in this domain reliable if someone want to scale this work up, or use it in another setting or environment.
			\item \textbf{Explicit Task}. The dataset should be set according to the problem to be solved, so that the relevant experimental results can verify the effectiveness of the model for solving this problem.
	\end{itemize}

\begin{figure}[!t]
	\centering
	\includegraphics[width=\linewidth]{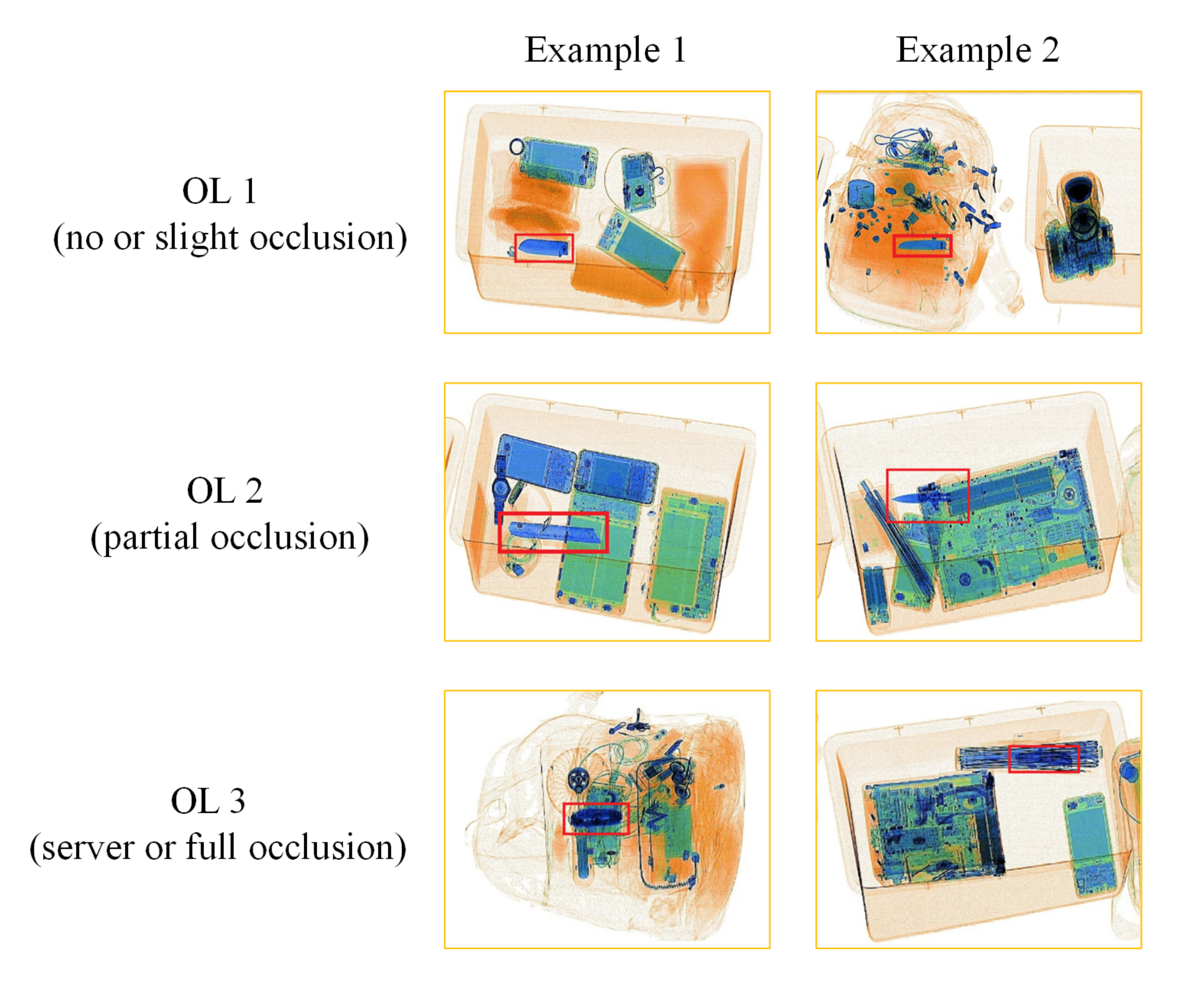}
	\caption{Samples of different occlusion levels in OPIXray dataset. ``OL'' refers to ``occlusion level''. There is no or slight occlusion on prohibited items in OL1, partial occlusion in OL2 and severe or full occlusion in OL3.}
	\label{Test_Occ}
\end{figure}

		Therefore, We construct our dataset based on the four principles. \textbf{First}, we select five categories of ``cutters'' as the common prohibited items to detect. The background images of the OPIXray dataset are generated by the latest X-ray imaging machine and gathered from daily security inspections in the international airport. \textbf{Second}, professional security inspectors synthesized the samples by embedding the specific prohibited item into the background images. And prohibited item in each sample is annotated manually with a bounding box by professional inspectors from the international airport, which is the procedure to train security inspectors. 
		\textbf{Third}, We followed the similar quality control procedures of annotation as the famous Pascal VOC. Each prohibited item is annotated by professional airport security inspectors. All inspectors followed the same annotation guidelines including what to annotate, how to annotate bounding, how to treat occlusion, etc. In addition, the accuracy of each annotation was checked by another professional inspector, including checking for omitted objects to ensure exhaustive labelling. \textbf{Finally}, for further study of the impact brought by object occlusion, we classify the testing set into three subsets according to the degree of occlusion, and name them in grades: Occlusion Level 1 (OL1), Occlusion Level 2 (OL2) and Occlusion Level 3 (OL3), where higher level implies severer occlusion around prohibited items in image. As illustrated in Fig. \ref{Test_Occ}, there is no or slight occlusion on prohibited items in OL1 and partial occlusion in OL2, and severe or full occlusion in OL3. Tab. \ref{data_cate_table} shows the category statistic of these three subsets with different occlusion levels.

	\subsection{Data properties}
	The OPIXray dataset contains a total of 8885 X-ray images, 5 categories of common cutters, namely, ``Folding Knife", ``Straight Knife", ``Scissor'', ``Utility Knife'' and ``Multi-tool Knife'' (illustrated in Fig. \ref{Samples_five}). Some images contain more than one prohibited item and every prohibited item is located with a bounding-box annotation. The statistics of category distribution are shown in Tab. \ref{data_table}. All images are stored in JPG format with the resolution of 1225*954. The dataset is partitioned into a training set and a testing set, where the ratio is about 4 : 1. The statistics of category distribution of training set and testing set are also shown in Tab. \ref{data_table}. Note that due to that some images contain more than one prohibited item, the sum of all items in the different categories is greater than the total number of images. According to our statistics, there are about 35 samples (30 in the training set and 5 in the testing set) containing more than one prohibited items.

	\subsection{Other Potential Tasks}
	%\subsection{Small X-ray Object Detection Dataset}
	In addition to occluded object detection, our OPIXray dataset can further serve the evaluation of various detection tasks including small object detection and few-shot object detection.
	
	\textbf{Small Object Detection}. Security inspectors often struggle to find small prohibited items in baggage or suitcase. In our OPIXray dataset, there are many small prohibited items. According to the definition of small by SPIE, the size of small object is usually no more than 0.12\% of entire image size. Suppose we define small object as the object whose max rectangle area is no more than 0.12\% of entire image, all the prohibited items in our dataset are small objects while other safe items (such as personal computers, umbrellas, etc.) are large items.

	\begin{table}[!t]
	\centering
	\renewcommand\arraystretch{1.2}
	\setlength{\tabcolsep}{9pt}
	\caption{The category distribution of different occlusion levels in the testing set of the OPIXray dataset.$^*$}
	\label{data_cate_table}
	\small
	\begin{threeparttable}
	\begin{tabular}{ccccccc}
		\hline
		\multirow{2}{*}{Testing} & \multicolumn{5}{c}{Categories}                     & \multirow{2}{*}{Total} \\ \cline{2-6}
		& FO & ST & SC & UN & MU &                        \\ \hline
		OL1                          & 206     & 88       & 160     & 214     & 255        & 922                    \\
		OL2                          & 148     & 84       & 126     & 88      & 105        & 548                    \\
		OL3                          & 50      & 63       & 83      & 41      & 70         & 306                    \\ \hline
		Total                        & 404     & 235      & 369     & 343     & 430        & 1776                   \\ \hline
	\end{tabular}
	\begin{tablenotes}
	\footnotesize
	\item $^*$Note that ``OL'' refers to ``Occlusion Level''.
\end{tablenotes}
\end{threeparttable}
\end{table}

	\textbf{Few-Shot Object Detection}. In the real security scene, The frequency of samples containing prohibited items is very low. As a result, images of the dataset can be used to construct a new dataset for few-shot object detection, which is more consistent with the real scene settings.
	
	Besides, in security inspection, some categories of prohibited items are extremely rare, leading to a very small amount of training data. Therefore, ``few-shot object detection'' can become a research track of X-ray prohibited items detection. Researchers can be inspired by relevant methods of few-shot learning or meta-learning to address the problem that some categories of prohibited items with few samples are hard to detect in security inspection.
	
	\begin{figure*}[!t]
	\centering
	\includegraphics[width=\linewidth]{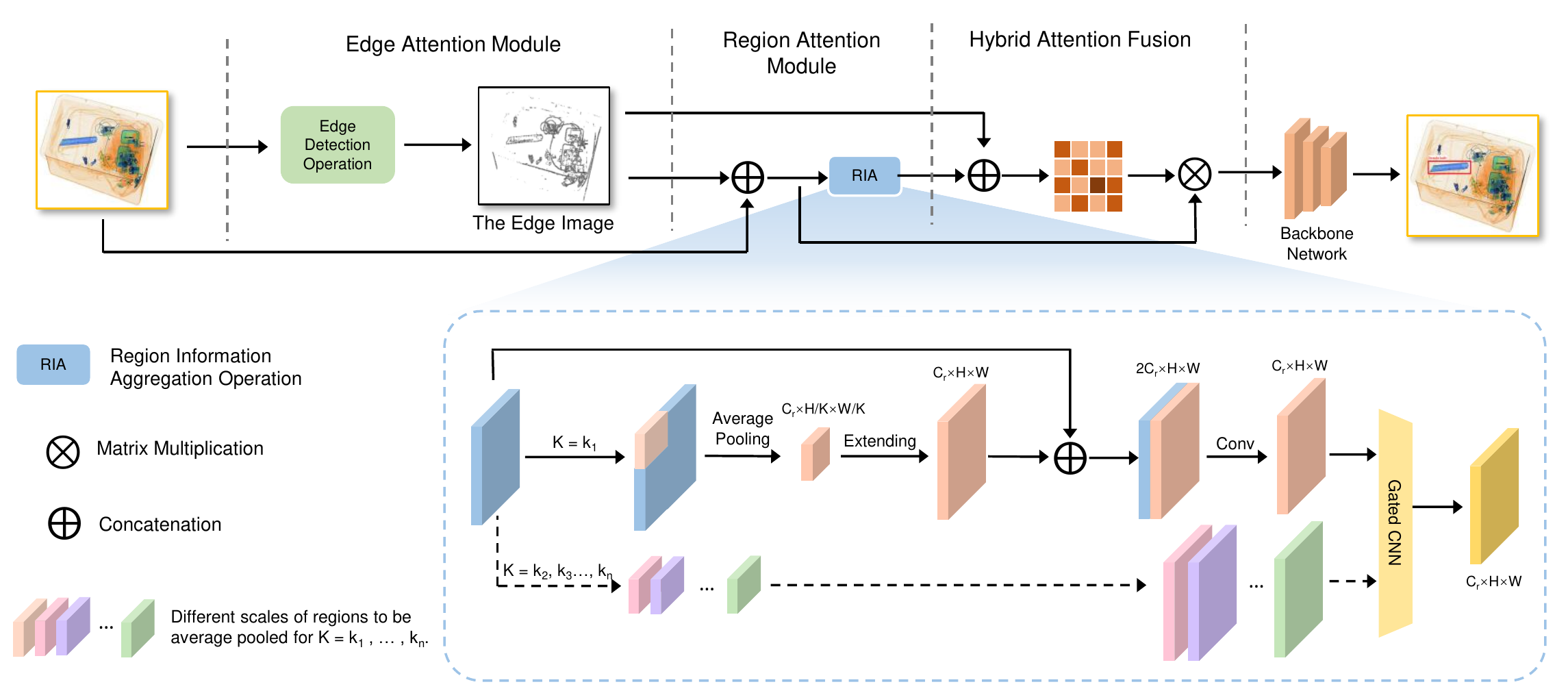}
	\caption{The network architecture of the de-occlusion attention module (DOAM) integrated with a general backbone network. As illustrated, two feature maps are generated by Edge Guidance (EG) and Material Awareness (MA) and fused to generate the attention map in Attention Generation. Further, the attention map is applied to the input image to generate refined feature maps we desire. Finally, the refined feature map can be utilized by the SSD network.}
	\label{doam_tmm}
\end{figure*}
	
	\section{Over-sampling de-occlusion attention network}\label{Section:model}
	This section describes the details of the proposed network, Over-sampling de-occlusion attention network (DOAM-O). As the de-occlusion attention module (DOAM) lies at the core of the proposed method, first, we elaborate the details of the de-occlusion attention module including the edge guidance (EG), material awareness (MA) and attention generation (AG). Second, we introduce the over-sampling strategy for selecting hard samples in detail. Finally, we discuss the essential difference between our method and other congeneric mechanisms.

	\begin{figure}[!t]
		\begin{algorithm}[H]
			\caption{The Procedure of DOAM.} \label{Alg}
			\begin{algorithmic}[1]
				\STATE \textbf{Input}: An X-ray image $\mathbf{x}\in\mathbb{R}^{C\times{H}\times{W}}$;
				\STATE Generate the horizontal edge image $\mathbf{E}^{h}$ and the vertical edge image $\mathbf{E}^{v}$ by the $Sobel$ operator.
				\STATE Generate the edge image ${E}$ by synthesizing $\mathbf{E}^{h}$ and $\mathbf{E}^{v}$.
				\FOR {$\mathrm{N}_1$ steps}
				\STATE Refine the feature map $\mathbf{A}$ through ${\mathcal{F}_\mathbf{e}}(\cdot)$.
				\ENDFOR
				\STATE Generate the image $\mathbf{P}$ by concatenating $\mathbf{x}$ and ${E}$.
				\FOR {$\mathrm{N}_2$ steps}
				\STATE Refine the feature map $\mathbf{B}_{1}$ through ${\mathcal{F}_\mathbf{m}}(\cdot)$.
				\ENDFOR
				\FOR {$k\in\left\{\mathbf{k}_1,\ldots,\mathbf{k}_n\right\}$}
				\STATE Generate the refined feature map $\mathbf{B}_{2}^k$ through Eq. (\ref{B_2}).
				\STATE Generate $\mathbf{B}_{3}^k$ by concatenating $\mathbf{B}_{1}$ and $\mathbf{B}_{2}^k$.
				\STATE Update the feature map set $\LARGE{U}$ = $\LARGE{U} \cup \mathbf{B}_{3}^k$.
				\ENDFOR
				\STATE Choose the appropriate feature map $\mathbf{B}$ from $\LARGE{U}$ by drawing the gated convolutional network $\mathcal{G}$.
				\STATE Generate the fused feature map $\mathbf{C}$ by operating $\mathbf{A}$ and $\mathbf{B}$.
				\STATE Generate the attention map $\mathcal{M}=\sigma(\mathbf{C})$.
				\STATE Generate the final feature map $\mathbf{D}$ by performing a matrix multiplication between $\mathcal{M}$ and $\mathbf{P}$ through Eq. (\ref{D}).
				\STATE \textbf{Output}: the final refined feature map $\mathbf{D}\in\mathbb{R}^{C_h\times{H}\times{W}}$.
			\end{algorithmic}
		\end{algorithm}
	\end{figure}

	\subsection{De-occlusion Attention Module}
	The DOAM simultaneously lays particular emphasis on edge information and material information of the prohibited item by utilizing two sub-modules, namely, Edge Guidance (EG) and Material Awareness (MA). Then, our module leverages the two information above to generate an attention distribution map as a high-quality mask for each input sample to generate high-quality feature maps, serving identifiable information for general detectors.
	\subsubsection{Edge Guidance (EG)}
	For each input sample $\mathbf{x}\in\LARGE{X}$, we compute the edge images $\textbf{E}^{h}$ and $\textbf{E}^{v}$ in horizontal and vertical directions using the convolutional neural network with $Sobel$ operators $s_h$, $s_v$, which denote the horizontal and vertical kernel respectively. By jointly exploit the above two results $\textbf{E}^{h}$ and $\textbf{E}^{v}$, the entire edge image $\textbf{E}$ of input image $\mathbf{x}$ is further generated.
	%we utilize the convolutional neural network with $s_h$, $s_v$ of the $Sobel$ operator, denoting the horizontal and vertical kernel, to respectively compute the edge images $\mathbf{E}^{h}$ and $\mathbf{E}^{v}$ in horizontal and vertical directions, respectively. We further generate the edge image $\mathbf{E}$ of the input image $\mathbf{x}$ by synthesizing the above two results $\mathbf{E}^{h}$ and $\mathbf{E}^{v}$.
	Next, we define $\mathrm{N}_1$ as the Module Operation Intensity of EG, which has a strong correlation with the performance of module. In order to limit EG to only amplify the edge information of the prohibited items, we use $\mathrm{N}_1$ network blocks, each of which consists of a convolutional layer with a $3\times3$ kernel size, a batch normalization layer, and a ReLU layer, to extract the feature map $\mathbf{A}$, which is the final output of the EG module.
	The operations can be formulated as follows:
	\begin{equation}
	\mathcal{F}_{\mathbf{e}}(\mathbf{x})=\mathrm{ReLU}\left(\mathbf{W}_{e}\cdot\mathbf{x}+\mathbf{b}_{e}\right),
	\end{equation}
	\begin{equation}\label{F:EG}
	\mathbf{A}=\left\{\mathcal{F}_{\mathbf{e}}({E})\right\}_{\mathrm{N}_1},
	\end{equation}
	where $\left\{\cdot\right\}_{\mathrm{N}_1}$ means that the operation is repeated $\mathrm{N}_1$ times. $\mathbf{W}_{e}$, $\mathbf{b}_{e}$ are parameters of the convolutional layer. As shown in Eq. \ref{F:EG}, the feature map $\mathbf{A}$ laying emphasis on edge guidance information is extracted, which helps the model to adaptively give more attention to the edge features of the prohibited items.
	\begin{figure*}[!t]
		\centering
		\includegraphics[width=\linewidth]{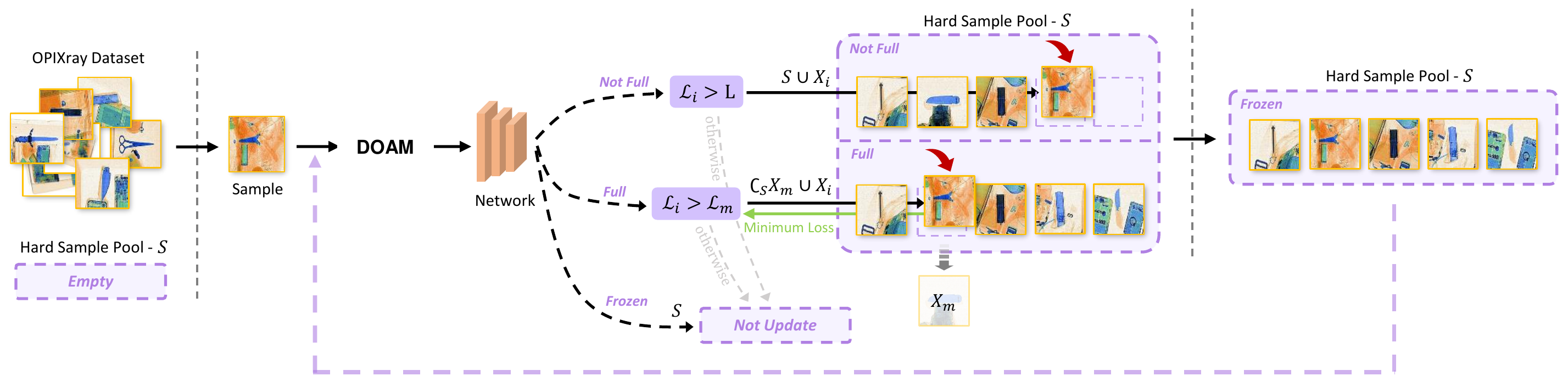}
		\caption{The over-sampling strategy of DOAM-O. $\LARGE{X}_i$ refers to the $i$-th epoch training data inputted,  $\LARGE{X}_{m}$ refers the images of one batch whose loss is minimum in the ``hard sample pool'' $S$ and $\mathcal{L}_{m}$ refers to their loss value. $\mathcal{L}_i$ refers to the total loss of the batch including regression loss and classification loss and $\mathrm{L}$ refers to the standard threshold (artificial setting).}
		\label{framework}
	\end{figure*}
	\subsubsection{Material Awareness (MA)}
	Material information is highly reflected in color and texture. In contract to the color information, which can directly represent itself at any position of the picture, texture information can only be recognized and utilized in conjunction with its surroundings.
	Based on the common knowledge that people identify the material of an object according to both its color and texture, we aggregate these two information as so-called aggregated regional information, which is regarded to represent the material information.
	%%%%%%%%%%%%%%%%%%%%%%%%%%%%%%%%%%%%·Ö¸îÏß%%%%%%%%%%%%%%%%%%%%%%%%%%%%%%%%%%%%%%%%%%%%%%%%%%%%%%%%%%
	To build relations between each position of the concatenated image (the input image $\mathbf{x}$  and its edge image ${E}$ in EG module) and a certain region around it, we utilize $\mathrm{N}_2$ network blocks which is defined as the Module Operation Intensity of MA module. Similar to $\mathrm{N}_1$ in EG, $\mathrm{N}_2$ contains a convolutional layer with the kernel size is $3\times3$, a batch normalization layer, and a ReLU layer, to extract a temporary feature map $\mathbf{B}_{1}$, which is a temperature feature map outputted by the network blocks of MA module. These operations can be formulated as follows:
	\begin{equation}
	\mathcal{F}_{\mathbf{m}}(\mathbf{x})=\mathrm{ReLU}\left(\mathbf{W}_{m}\cdot\mathbf{x}+\mathbf{b}_{m}\right),
	\end{equation}
	\begin{equation}\label{F:EG}
	\mathbf{B}_1=\left\{\mathcal{F}_{\mathbf{e}}(\mathbf{x}||{E})\right\}_{\mathrm{N}_2},
	\end{equation}
	where $||$ denotes concatenating operation and $\left\{\cdot\right\}_{\mathrm{N}_2}$ means that the operation is repeated $\mathrm{N}_2$ times. Moreover, we refine $\mathbf{B}_1$ through Region Information Aggregation (RIA) and further generate the refined feature map $\mathbf{B}$ as the final output of MA module.
	
	As shown in Fig. \ref{doam_tmm}, RIA operation takes two inputs, specifically the feature map $\mathbf{B}_{1}$ and a parameter $k$, and aggregates them by average pooling and extending operations
	with a fixed region size of $k\times k$, generating another temperate feature map $\mathbf{B}_{2}^k$.
	The average pooling and extending operations can be formulated together as follows:
	\begin{equation}\label{B_2}
	{\mathbf{B}_{2}}^k_{ij}=\frac{\sum _{m=i-(i\bmod{k})}^{i-(i\bmod{k})+k}\sum _{n=j-(j\bmod{k})}^{j-(j\bmod{k})+k}{\mathbf{B}_{1}}_{mn} }{k^2},
	\end{equation}
	where ``mod'' refers to the modulus operation and ${\mathbf{B}_{2}}^k_{ij}$ refers the feature of the $i$-th row and $j$-th column of feature map $\mathbf{B}_{2}$ when the kernel size for the average pooling layer is $k$.
	
	We further concatenate the two feature maps ($\mathbf{B}_{1}$ and $\mathbf{B}_{2}$) in the dimension of channel to generate a new feature map $\mathbf{B}_{3}$ with the dimentions of $2C_r\times{H}\times{W}$, in which every point is able to perceive a certain region around it with size of $k\times k$.
	In order to be compatible with various region size (due to different values of $k$), the module generates a set of feature maps, $\LARGE{U}$=$\left\{\mathbf{B}_{3}^{k_1},\cdots,\mathbf{B}_{3}^{k_n}\right\}$.
	
	Since the scales of prohibited items vary a lot, we design a mechanism to adaptively choose an optimal value of which enables RIA operation to perform well in most situations.
	We integrate the gated convolutional neural network \cite{yu2019free} $\mathcal{G}$ with $3\times3$ kernels into RIA, to select the proper feature map $\mathbf{B}$ from the set $\LARGE{U}$ as output. The operations can be formulated as follows:
	\begin{equation}\label{B}
	\mathbf{B} = \mathcal{G}(U),
	\end{equation}
	where $U$=$\left\{\mathbf{B}_{3}^{k_1},\cdots,\mathbf{B}_{3}^{k_n}\right\}$.
	
	\subsubsection{Attention Generation}
	
	As is illustrated in Algorithm \ref{Alg}, feature map $\mathbf{A}\in \mathbb{R}^{C_e\times{H}\times{W}}$ is outputted by EG module, and $\mathbf{B}\in \mathbb{R}^{C_r\times{H}\times{W}}$ is choosen from a refined feature map set $\LARGE{U}$ in the MA module. Next,in order to fuse the two knids of information, we concatenate the two feature maps genertated by the two core sub-modules and feed them into a convolutional layer with 1$\times$1 kernel size. Therefore, a fused feature map is then generated, $\mathbf{C}\in \mathbb{R}^{(C_e+C_r)\times{H}\times{W}}$, which integrates edge and material information which have been strengthened.
	These operation can be formulated as follows:
	\begin{equation}\label{F:mix}
	\mathbf{C}=\mathbf{W}_{a}\left(\mathbf{A}||\mathbf{B}\right)+\mathbf{b}_{a},
	\end{equation}
	where $||$ represents the operation of concatenating, and $\mathbf{W}_{a}$, $\mathbf{b}_{a}$ are parameters of the convolutional layer. Then we utilize the feature map $\mathbf{C}$ as the input of a sigmoid function to generate the attention map $\mathcal{M}$:
	\begin{equation}\label{M}
	\mathcal{M}=\sigma(\mathbf{C}) = \frac{1}{1+e^{-\mathbf{C}}},
	\end{equation}
	where $\mathcal{M}\in\mathbb{R}^{{H}\times{W}}$. Finally, we calculate the inner product of the attention map $\mathcal{M}$ and the concatenated image $\mathbf{P}$ to acquire the final refined feature map $\mathbf{D}$:
	\begin{equation}\label{D}
	\mathbf{D}_{j}=\sum_{i=1}^{H\times{W}}\mathcal{M}_{ji}\mathbf{P}_i,
	\end{equation}
	where $\mathbf{D}\in\mathbb{R}^{{C}_{h}\times{H}\times{W}}$ and $\mathbf{P}$ refers to the concatenated image of the input image $\mathbf{x}$ and its corresponding edge image ${E}$. The final feature map $\mathbf{D}$ emphasizes the information of the prohibited items and therefore highly contributes to the detection if served to detectors.

	\begin{figure}[!t]
	\begin{algorithm}[H]
		\caption{The Procedure of the over-sampling strategy.} \label{Alg}
		\begin{algorithmic}[1]
			\STATE \textbf{Input}: The sample batches $\LARGE{X}_1,\LARGE{X}_2,\dots,\LARGE{X}_\mathcal{N}$, the standard loss threshold $\mathrm{L}$, the hard sample pool $S$ with size $N_S$;
			\FOR {$i = 0 \to \mathcal{N}$}
			\STATE Calculate the loss $\mathcal{L}_i$ through Eq. \ref{loss_i}.
			\IF {$S\,$ isnot$\, full$}
			\IF {$\mathcal{L}_{i}>\mathrm{L}$}
			%\STATE $S = \complement_{S}\LARGE{X}_{m} \cup\LARGE{X}_{i}$.
			\STATE $S=S \cup X_{i}$.
			\ENDIF
			\ELSE
			\STATE Find the batch ${X}_{m}$ having minimum loss $\mathcal{L}_{m}$ in $S$.
			\IF {$\mathcal{L}_{i}>\mathcal{L}_{m}$}
			\STATE $S = \complement_{S}\LARGE{X}_{m} \cup\LARGE{X}_{i}$.
			\ENDIF
			\ENDIF
			\ENDFOR
			\STATE \textbf{Output}: The hard sample pool $S=\{X_{1},X_{2},\dots,X_{N_S}\}$.%\left\{X_{1},X_{2},\dots,X_{N_S}\}\right$.
		\end{algorithmic}
	\end{algorithm}
\end{figure}

	\subsection{Over-sampling Strategy}
	Suppose we select $n$ samples from the whole training set for each batch and the size of ``hard sample pool'' $S$ is $N_S$. As illustrated in Fig. \ref{framework}, first, we input the $i$-th epoch training data, $\LARGE{X}_i = \left\{\mathbf{x}_1,\cdots,\mathbf{x}_n\right\}$, into the backbone network to extract features. Second, we calculate the total loss of each batch including regression loss and classification loss. The loss function can be formulated as follows:
	\begin{equation}
	\mathcal{L}_i=\frac{\sum_{i=1}^{n}\left(\mathcal{L}_{loc}+\mathcal{L}_{\text {conf }}\right)}{n}.
	\label{loss_i}
	\end{equation}
	Third, we compare $\mathcal{L}_i$ with the standard threshold $\mathrm{L}$ (artificial setting). In the case of that $\mathcal{L}_i$ is smaller than $\mathrm{L}$, the gradient calculation and back propagation are carried out directly and this training iteration is exited. In the case of that $\mathcal{L}_i$ is larger than $\mathrm{L}$, if the number of batches in $S$ is smaller than $N_S$, we regard the images of this batch as ``hard samples'' and put them into $S$. The hard sample selection function $\mathcal{F}_S$ can be regareded as the update process of ``hard samples pool'' $S$. Therefore, in the case of the hard sample pool is not full, the selection function can be formulated as follows:
	\begin{equation}
	\mathcal{F}_S: \quad S=\left\{\begin{array}{ll}
	S \cup X_{i} & , \quad \mathcal{L}_{i}>\mathrm{L} \\
	S & , \text { otherwise }
	
	\end{array}. \right.
	\end{equation}
	If the number of batches in $S$ is larger than $N_S$, we compare $\mathcal{L}_i$ with the minimum loss value of these batch in $S$ already. In the case of that $\mathcal{L}_i$ is larger than the minimum loss value in $S$, we update $S$ by replacing the batch images whose loss is minimum with this batch images. In the case of that $\mathcal{L}_i$ is smaller than the minimum loss value in $S$, this training iteration is exited. Therefore, in the case of the hard sample pool is full, the selection function can be formulated as follows:
	\begin{equation}
	\mathcal{F}_S: \quad S=\left\{\begin{array}{ll}
	\complement_{S}\LARGE{X}_{m} \cup\LARGE{X}_{i} & , \quad \mathcal{L}_{i}>\mathcal{L}_{m} \\
	S & , \quad \text { otherwise }
	\end{array}, \right.
	\end{equation}
	where $\LARGE{X}_{m}$ means the images of one batch whose loss is minimum in the ``hard sample pool'' $S$ and $\mathcal{L}_{m}$ refers to their loss value.
	Finally, after this training iteration is finished, a ``hard sample pool'' $S$ with $N_S$ batches ``hard samples'' is generated. The images in $S$ are retrained to help the model learn more detailed information from ``difficlties''. After these ``hard samples'' are retrained, we clear $S$ and start the next training iteration. By exploiting this over-sampling strategy to retrain these hard samples, the model reduced the performance decrease caused by hard samples in the training set significantly and enhanced the ability of generalization highly, just as that human learns a lot from the difficulties they felt hard to solve.

%We propose the De-occlusion Attention Module(DOAM), which utilizes two sub-modules, named Edge Guidance (EG) and Material Awareness (MA), to simultaneously laying particular emphasis on edge information of the occluded part and material information of the visual part of the prohibited item.
%Then, these two types of informations are exploited to generate an attention distribution map as a high-quality mask for the input sample, and the feature maps are then enhanced by the attention maps to serve identifiable information for general detectors.
%To generate enhanced feature maps, serving identifiable information for general detectors, DOAM further leverages the two information above to generate an attention distribution map as a specific mask for the input sample.
%We demonstrate our design from the following aspects: 1) how DOAM work briefly (A); 2) how to extract the edge information (B); 3) how to aggregate the material information (C); 4) how to generate the attention distribution map (D); 5) How the performance of the module changes with the module operation intensity (E); 6) how to compare our module with the base detector and other counterparts (F).

%The entire process of DOAM is illustrated in detail in Algorithm \ref{Alg}.
	
\subsection{Comparison with other attention mechanisms of DOAM}

\textbf{Non-local \cite{wang2018non}.} Non-local network captures the long dependence on the feature map by establishing the relationship between each pixel on the feature map and the global pixels, so that the network can better predict the location of the target according to the relationship between the objects in the image. As \cite{wang2018non} introduced, the non-local attention mechanism can be formulated as follows:
\begin{equation}
\text{Non-local}:\quad y_{i}=\frac{1}{C(x)} \sum_{\forall j} f\left(x_{i}, x_{j}\right) g\left(x_{j}\right),
\end{equation}
where $j$ refers to all other positions except $i$ in the feature map. However, prohibited items in X-ray images are often small in size, and has little relations with long-distance areas filled of safety items. Too much consideration of the relationship between each position and all other positions on the feature map will distract the attention of the network, causing the performance not satisfactory. Therefore, we narrow down the scope of considering local relations and artificially set several receptive fields and let the gate convolution choose the most appropriate range. This strategy of RIA module of DOAM in Fig. \ref{framework} can be considered as a variant of non-local mechanism.
\begin{equation}
\text{DOAM}:\quad y_{i}=\frac{1}{k^{2}} \sum_{\forall j \in R_{k}}\left(x_{i}+x_{j}\right),
\end{equation}
where $R_{k}$ refers to the range of distance $k$ centered on $i$.

\textbf{SE \cite{hu2018squeeze}.} SENet focuses on assigning different attention weights to the feature maps of each channel, and then optimizes the channel feature maps which have great influence on the prediction results. This is essentially to make the model pay more attention to the channel feature maps containing key features. As \cite{hu2018squeeze} introduced, this attention mechanism can be formulated as follows:
\begin{equation}
\text{SE}:\quad \mathbf{s}=\mathbf{F}(\mathbf{z}, \mathbf{W}),
\end{equation}
where $\mathbf{z}$ refers to the set consisting of all channel feature maps.
By analyzing the characteristics of the security image, the edge features of the prohibited items have an important influence on the prediction performance of the model. Therefore, we directly use the edge image (illustrated in Fig. \ref{framework}) to generate the attention map to leading the model to focus on the edge features. In addition, we concatenate the edge image as an individual channel with other feature maps, which can be considered as a variant of SE mechanism that one channel are artificially set to edge image and this mechanism can be formulated as follows:
\begin{equation}
\text{DOAM}:\quad\mathbf{s}=\mathbf{F}(\mathbf{z} \cup z_e, \mathbf{W}),
\end{equation}
where $z_e$ refers to the edge image of the original input images in Fig. \ref{framework}.
Compared with SE, we integrate prior knowledge into the training process of the model, which can achieve better results for specific problems.

\textbf{DA \cite{fu2019dual}.} DA takes into account the relationship between location and channel on the feature map, and adopts the hybrid attention mechanism of space and channel. Similar to the location attention mechanism in non-local, DA construct the relations between each position and all of other position in the whole feature maps. Besides, the channel attention mechanism of DA is similar to SE that modeling long distance dependence between different channels. Inspired by this mechanism, we fuse the features outputted by edge guidance module and material awareness module, and further generate an attention map considering the edge and material characteristics comprehensively.

\subsection{Comparison with other congeners of sampling strategy}

\textbf{Easy sample over-sampling strategy}. One strategy is simple resampling, which restricts the network to focus more attention on the sample data that is easy to identify. Empirically, this strategy sometimes ensures the location and category of dangerous goods with a higher degree of confidence can be more accurately predicted, and further reduces the false recognition of dangerous goods areas in the security inspection pictures. Similar to the definition above, in the case of the sample pool is not full, the selection function can be formulated as follows:
\begin{equation}
\mathcal{F}_S:\quad S=\left\{\begin{array}{ll}
S \cup X_{i} & , \quad \mathcal{L}_{i}<\mathrm{L} \\
S & , \text { otherwise }
\end{array}.\right.
\end{equation}

Besides, in the case of the sample pool is full, the selection function can be formulated as follows:
\begin{equation}
\mathcal{F}_S:\quad S=\left\{\begin{array}{ll}
\complement_{S}\LARGE{X}_{m} \cup\LARGE{X}_{i} & , \quad \mathcal{L}_{i}<\mathcal{L}_{max} \\
S & , \quad \text { otherwise }
\end{array}, \right.
\end{equation}
where $\mathcal{L}_{max}$ refers to the loss value of the batch whose loss is maximal in $S$ and other variables are the same as the definitions above.

However, the usage of this training strategy will miss more difficult sample data, which can not guarantee the principle that the dangerous goods in the security pictures should be checked as much as possible.

\textbf{Random sample over-sampling strategy}. Another strategy is enhancing the generalization ability of the model and reduce the biased prediction of the model by introducing random sample data. The selection function can be formulated as follows:
\begin{equation}
\mathcal{F}_S:\quad S=\left\{X_{1}, X_{2},\dots, X_{N_S}\},\right.
\end{equation}
where $X_{1}, X_{2},\dots, X_{N_S}$ are the batches randomly selected from the training data.
However, random sample over-sampling strategy has no clear purpose and suffers more uncertainty compared to the hard sample resampling method, thereby it is also not a satisfactory training strategy.
	
\textbf{Focal Loss \cite{lin2017focal}}. The Focal Loss strategy uses weighted cross-entropy loss and adaptively enhance the learning ability of hard samples. The loss function of this method is as follows:
\begin{equation}
\mathcal{L}\left(p_{\mathrm{t}}\right)=-\left(1-p_{\mathrm{t}}\right)^{\gamma} \log \left(p_{\mathrm{t}}\right),
\end{equation}
where $p_{\mathrm{t}} \in \left[0,1\right]$ and ${\gamma} \ge 0$. Note that this learning method mainly focuses on the hard samples in the training data, and its essence is the same as our hard sample resampling method. However, due to the existence of some outliers in the training data, paying more attention to these extremely hard samples will cause the model to ignore the correct optimization direction to fit these samples. Fortunately, the hard sample resampling method gives similar weights to the data in the sample pool, which will not lead to over learning of outliers.

	\section{Experiments}\label{Section:exp}
	In this section, extensive experiments are conducted to evaluate the model we proposed. In our work, the primary goal is to detect occluded prohibited items in X-ray images in security inspection scenario. As far as we know, no dataset focusing on this task has been proposed up to now, so we only employ the OPIXray dataset for evaluation all through the following experiments. \textbf{First}, we demonstrate that DOAM-O outperforms all the attention mechanisms mentioned above, over different categories and distinctive occlusion levels. \textbf{Second}, we perform ablation experiments to completely evaluate the effectiveness of DOAM-O. \textbf{Third}, we verify the overall compatibility and effectiveness of DOAM-O after integrating with various object detection networks. \textbf{Finally}, we visualize the generated attention distribution map and the whole network to give a more intuitive verification.
	
	\textbf{Evaluation strategy:} All experiments are proceeded on the OPIXray dataset and trained on training set in Tab. \ref{data_table}. In most experiments, models are tested by the testing set data in Tab. \ref{data_table}. As for comparing different attention mechanisms in terms of occlusion levels, models are tested on OL1, OL2 and OL3 in Tab. \ref{data_cate_table} respectively.
	
	\textbf{Baseline Detail:} To make fair comparisons between different attention mechanisms, we plug DOAM and each of the other attention modules into SSD respectively, and evaluate the performance of these integrated networks as well as SSD itself.
	These attention modules are added to the backbone (VGG16) of SSD. More explicitly, they are embedded behind the max pooling layer where the feature map is refined to half. In ablation study, we plug each sub-module of DOAM into SSD \cite{liu2016ssd} one by one and report the performance of SSD \cite{liu2016ssd} and report the results under different conditions to evaluate the utility of various sub-modules. Lastly, we evaluate the compatibility of DOAM by integrating it into several mainstream detection networks, including YOLOv3, FCOS, as well as SSD.
	
	\textbf{Parameter setting:} All through the experiments, models are optimized using SGD optimizer with initial learning rate at 0.0001. The batch size is set to 24 and the momentum and weight decay are set to 0.9 and 0.0005 respectively. We utilize the mean Average Precision (mAP) as the metric to evaluate the model performance and the IOU threshold is set to 0.5. We calculate the AP of each category using the model with the  best performance to observe the improvement in different categories. Moreover, to avoid the bias from data transformation while generating edge image, no data augmentation technique are used to modify the pixel value of the original image, which leads to a better analysis of the impact of edge information.
	
	\subsection{Comparing with Different Attention Mechanisms}
	We compare three variants of attention mechanisms, including SE \cite{hu2018squeeze}, Non-local \cite{wang2018non} and DA \cite{fu2019dual}. Tab. \ref{table_attention_catagories} and \ref{Occlusion-result} reports the performances of all models. Note that ``FO'', ``ST'', ``SC'', ``UT'' and ``MU'' refer to ``Folding Knife'', ``Straight Knife'', ``Scissor'', ``Utility Knife'' and ``Multi-tool Knife'', respectively. Besides, ``DOAM-T'' refers to the model ``SSD+DOAM'' with traditioanl training strategy and ``DOAM-O'' refers to the model ``SSD+DOAM'' with over-sampling training strategy.
	
	\subsubsection{Object Categories}
	As is shown in Tab. \ref{table_attention_catagories}, DOAM not only improves the performance of SSD by $3.12\%$, but also outstrips SE, Non-local, and DA methods by
	$2.16\%$, $2.60\%$, $2.05\%$, respectively. Especially in the category of Straight Knife, Folding Knife and Utility Knife, DOAM achieves a large improvemnt.
	
	Particularly for Straight Knife, which is the category with the worst occlusion, DOAM sees a remarkable amount over SSD \cite{liu2016ssd} baseline by $6.48\%$ and Non-local \cite{wang2018non} by $5.12\%$. The performance of DOAM in Scissor, the category with the lightest occlusion, is merely improved by $1.71\%$ compared with SSD \cite{liu2016ssd} and similar to Non-local \cite{wang2018non}. Obviously, DOAM surpasses these current widely-used attention mechanisms over different categories in the field of prohibited item detection based on X-ray images.
	
	\begin{table}[h]
		\centering
		\renewcommand\arraystretch{1.3}
		\setlength{\tabcolsep}{5pt}
		\caption{Performance comparison between DOAM-O and other different attention mechanisms on object categories.}
		\label{Classification-result}
		\small
		\begin{tabular}{lcccccc}
			\hline
			\multicolumn{1}{l}{\multirow{2}{*}{Method}} & \multicolumn{1}{c}{\multirow{2}{*}{mAP}} & \multicolumn{5}{c}{Categories}                                                        \\ \cline{3-7}
			\multicolumn{1}{l}{}
			& \multicolumn{1}{c}{}
			& \multicolumn{1}{c}{FO} & \multicolumn{1}{c}{ST} & \multicolumn{1}{c}{SC} & \multicolumn{1}{c}{UT} & \multicolumn{1}{c}{MU} \\ \hline
			SSD \cite{liu2016ssd} & 70.89  & 76.91  & 35.02  & 93.41    & 65.87& 83.27\\
			\hline
			+SE \cite{hu2018squeeze}  & 71.85  & 77.17 & 38.29  & 92.03  & 66.10   & \textbf{85.67} \\
			+Non-local \cite{wang2018non}  & 71.41 & 77.55 & 36.38    & 95.26   & 64.86   & 82.98 \\
			+DA \cite{fu2019dual} & 71.96   & 79.68 & 37.69    & 93.38  & 64.14  & 84.90\\
			+DOAM-T &74.01&81.37&41.50&95.12&68.21&83.83\\
			\textbf{+DOAM-O} &\textbf{74.57} & \textbf{81.67} & \textbf{41.90} & \textbf{95.37} & \textbf{68.41} & 84.23\\
			\hline
		\end{tabular}\label{table_attention_catagories}
	\end{table}
	
	\subsubsection{Object Occlusion Levels}
\begin{figure}[!t]
		\centering
		\includegraphics[width=0.9\linewidth]{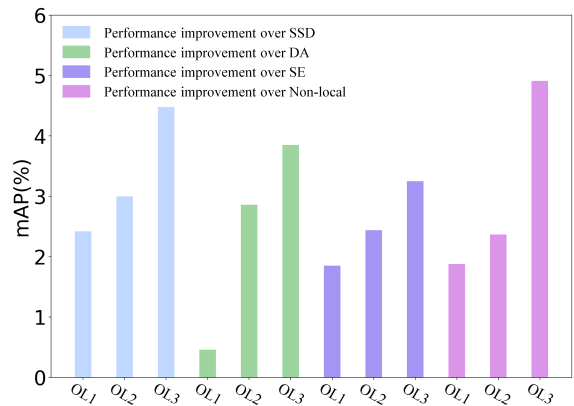}
		\caption{The amount changes of performance improvement of DOAM over different models with occlusion level increasing.}
		\label{improve}
	\end{figure}

	Tab. \ref{level-table} shows different attention mechanisms with SSD on each object occlusion level. And Fig. \ref{improve} is drawning from Tab. \ref{level-table}, which illustrates the improvement of DOAM compared with other methods in terms of different occlusion levels.
	
	In other words, with the increasing of occlusion level, the model gets improved more obviously. It implies that DOAM is able to tackle the occlusion issue especially in heavily occluded cases.
	(Note that in OL3, the performance of ``SSD+Non-local" is lower than ``SSD". We conjecture the decrease is closely related to the attention mechanism of Non-local, which captures spatial information by building the relations between regions. Unfortunately, this type of relation reduces effect with the increasingly disorder of image.)
	
	\begin{table}[h]
		\centering
		\renewcommand\arraystretch{1.3}
		\setlength{\tabcolsep}{12pt}
		\caption{Performance comparison between DOAM and other different attention mechanisms on object occlusion levels.}
		\label{Occlusion-result}
		\small
		\begin{tabular}{lccc}
			\hline
			Method & OL 1 & OL 2 & OL 3\\ \hline
			SSD \cite{liu2016ssd} & 75.45 & 69.54 & 66.30\\
			\hline
			+SE \cite{hu2018squeeze} & 76.02 & 70.11 & 67.53\\
			+Non-local \cite{wang2018non} & 75.99 & 70.17 & 65.87\\
			+DA \cite{fu2019dual} & 77.41 & 69.68 & 66.93\\
			\textbf{+DOAM} & \textbf{77.87} & \textbf{72.45} & \textbf{70.78}\\
			\hline
		\end{tabular}
		\label{level-table}
	\end{table}
	
	\subsection{Comparing with Different Detection Approaches}
	To further evaluate the effectiveness and applicability of DOAM, we conduct experiments on three popular detection approaches, SSD \cite{liu2016ssd}, YOLOv3 \cite{redmon2018yolov3} and FCOS \cite{tian2019fcos}. The results are shown in Tab. \ref{detection_approach}.
	\begin{table}[!h]
		\centering
		\renewcommand\arraystretch{1.3}
		\setlength{\tabcolsep}{5pt}
		\caption{Performance comparison between DOAM-integrated network and baselines for three famous detection approaches.}
		\small
		\begin{tabular}{lcccccc}
			\hline
			\multicolumn{1}{l}{\multirow{2}{*}{Method}} & \multicolumn{1}{c}{\multirow{2}{*}{mAP}} & \multicolumn{5}{c}{Category} \\ \cline{3-7}
			\multicolumn{1}{l}{}
			& \multicolumn{1}{c}{}
			& \multicolumn{1}{c}{FO} & \multicolumn{1}{c}{ST} & \multicolumn{1}{c}{SC} & \multicolumn{1}{c}{UT} & \multicolumn{1}{c}{MU} \\ \hline
			SSD \cite{liu2016ssd} & 70.89  & 76.91  & 35.02  & 93.41    & 65.87& 83.27 \\
			+DOAM-T  & 74.01  & \textbf{81.37} & 41.50  & 95.12  & 68.21  & 83.83 \\
			\textbf{+DOAM-O} &\textbf{74.57} & \textbf{81.67} & \textbf{41.90} & \textbf{95.37} & \textbf{68.41} & \textbf{84.23}\\
			\hline
			YOLOv3 \cite{redmon2018yolov3}  & 78.21 & \textbf{92.53} & 36.02    & \textbf{97.34}   & 70.81   & 94.37 \\
			+DOAM-T   & 79.25   & 90.23 & 41.73   & 96.96  & 72.12  & 95.23\\
			\textbf{+DOAM-O}   & \textbf{80.40}   & \textbf{92.00} & \textbf{43.57}    & \textbf{97.70}  & \textbf{73.05}  & \textbf{95.65}\\
			\hline
			FCOS \cite{tian2019fcos} & 82.02 & 86.41 & 68.47 & 90.22 & 78.39 & 86.60\\
			+DOAM-T & 82.41 & 86.71 & 68.58 & \textbf{90.23} & 78.84 & 87.67\\
			\textbf{+DOAM-O} & \textbf{83.80} & \textbf{87.61} & \textbf{72.73} & 90.03 & \textbf{80.77} & \textbf{87.80}\\
			\hline
		\end{tabular}\label{detection_approach}
	\end{table}
	
	As shown in Tab. \ref{detection_approach}, the performance of DOAM-integrated networks are promoted by 3.12\%, 1.04\% and 0.39\% in comparison with SSD \cite{liu2016ssd}, YOLOv3 \cite{redmon2018yolov3} and FCOS \cite{tian2019fcos} respectively, which means that our module can be utilized as a plug-and-play module into most detection networks. Note that the performances on Folding Knife and Scissor are slightly diminished after DOAM-integrated.
	
	We speculate it is because our attention mechanism focuses more on the parts occludede heavily, so it is better at dealing with high-level occlusions. While samples in these two categories above are slightly occluded, which may lead to the little fall in performance for Folding Knife and Scissor.
	
	\subsection{Comparing with Other Over-sampling Mechanisms}
	In this section, to verify the effectiveness of our over-sampling strategy, we compare three variants of retraining strategies, including easy-sample retraining strategy (+Easy), random-sample retraining strategy (+Random) and Focal Loss strategy (+Focal). Tab. \ref{oversamplingstr} reports the performances of all models. Note that in Tab. \ref{oversamplingstr}, the four training strategies are implemented in the base model (SSD+DOAM) for fair comparison. Besides, ``DOAM-T'' refers to the model ``SSD+DOAM'' with traditioanl training strategy and ``DOAM-O'' refers to the model ``SSD+DOAM'' with the over-sampling training strategy.
	
	\begin{table}[h]
		\centering
		\renewcommand\arraystretch{1.3}
		\setlength{\tabcolsep}{5pt}
		\caption{Performance comparison between the over-sampling strategy and other congeneric mechanisms on object categories.$^*$}
		%\caption{{\upshape Note that ``DOAM-E'' refers to the model ``SSD+DOAM'' with easy sample over-sampling training strategy.}}
		\label{Classification-result}
		\small
		\begin{threeparttable}
		\begin{tabular}{lcccccc}
			\hline
			\multicolumn{1}{l}{\multirow{2}{*}{Method}} & \multicolumn{1}{c}{\multirow{2}{*}{mAP}} & \multicolumn{5}{c}{Categories}                                                        \\ \cline{3-7}
			\multicolumn{1}{l}{}
			& \multicolumn{1}{c}{}
			& \multicolumn{1}{c}{FO} & \multicolumn{1}{c}{ST} & \multicolumn{1}{c}{SC} & \multicolumn{1}{c}{UT} & \multicolumn{1}{c}{MU} \\ \hline
			SSD \cite{liu2016ssd} & 70.89  & 76.91  & 35.02  & 93.41    & 65.87& 83.27\\
			+DOAM-T &74.01&81.37&41.50&95.12&68.21&83.83\\
			\hline
			+DOAM-E  & 73.22  & 81.36 & 36.67  & 92.38  & 71.21   & \textbf{84.47} \\
			+DOAM-R  & 73.42 & 81.28 & 35.57    & 95.21   & 71.89   & 83.12 \\
			+DOAM-F \cite{fu2019dual} & 73.25   & 80.69 & 34.59    & 95.25  & \textbf{73.34}  & 82.40\\
			\textbf{+DOAM-O} &\textbf{74.57} & \textbf{81.67} & \textbf{41.90} & \textbf{95.37} & 68.41 & 84.23\\
			\hline
		\end{tabular}\label{oversamplingstr}
		\begin{tablenotes}
			\footnotesize
			\item $^*$Note that ``E'' refers to the easy sample over-sampling training strategy, ``R'' refers to the random sample over-sampling training strategy and ``F'' refers to the focal loss strategy.
		\end{tablenotes}
		\end{threeparttable}
	\end{table}
	
	\subsection{Module Complexity Analysis}
	In this section, we analyze the model complexity including the total number of parameters, model size and computation cost of different attention mechanisms in the case of that the base model is SSD. The three attention mechanisms that we make comparison with DOAM are SE \cite{hu2018squeeze}, Non-local \cite{wang2018non} and DA \cite{fu2019dual}, which focus on channel information, spatial information, and combination of the two kinds of information above respectively. All of them lead to quite a bit of extra parameter size and model size, while DOAM only brings negligible addition (about 0.41\% and 0.11\% in parameter size and model size) to SSD without attention mechanisms. As for computational cost, our DOAM slightly brings extra computational cost compared with SSD without any attention mechanisms (about 7.14\% in GFLOPs), meanwhile SE, Non-local, and DA bring 3.15\%, 6.54\% and 23.07\% respectively.
	\begin{table}[h]
		\centering
		\renewcommand\arraystretch{1.3}
		\setlength{\tabcolsep}{8pt}
		\caption{Complexity comparison of different models.$^*$}
		\label{param-result}
		\small
		\begin{threeparttable}
		\begin{tabular}{lccc}
			\hline
			Method & PARAMs & SIZE(MB) & GFLOPs\\
			\hline
			SSD \cite{liu2016ssd} & $24.2\times10^6$ & 92.6 & 30.6522\\
			\hline
			+SE \cite{hu2018squeeze}  &   $32.0\times10^6$  &  122.1 & \textbf{31.6169}\\
			+Non-local \cite{wang2018non}  & $30.9\times10^6$ & 117.9 &  32.6577\\
			+DA \cite{fu2019dual} & $45.6\times10^6$ & 174.1 & 37.7231\\
			\textbf{+DOAM} & $\bm{24.3\times10^6}$ & \textbf{92.7} & 32.8435\\
			\hline
		\end{tabular}
		\label{param-table}
				\begin{tablenotes}
			\footnotesize
			\item $^*$Note that ``PARAMs'', ``SIZE'' and ``GFLOPs'' refer to ``the total number of parameters'', ``the Model Size'' and ``the Giga Floating Point operations'', respectively.
		\end{tablenotes}
	\end{threeparttable}
	\end{table}
	
	DOAM prevails the other three attention mechanisms in parameter size and model size, while is slightly more expensive in computational cost. It might result from different values of parameter which causes repetitive computation in RIA.
	
	\subsection{Ablation Study}\label{sec:ablation-study}
	In this section, we report the results of ablation studies in Tab. \ref{Ablation-result} to verify the effectiveness of each part of our model. Note that in Tab. \ref{Ablation-result}, "C" refers simply concatenating the edge image and original RGB image. "DOAM/MA" refers DOAM without Material Awareness module and "DOAM/$\mathcal{G}$" refers DOAM without the Gate Convolutional Neural Network. "DOAM" refers to the DOAM with traditional training strategy and ``DOAM-O'' refers to the model we proposed in this paper.

	\begin{table}[!h]
	\centering
	\renewcommand\arraystretch{1.3}
	\setlength{\tabcolsep}{5pt}
	\caption{Ablation studies of Over-sampling De-occlusion Attention Network.$^*$}
	\small
	\begin{threeparttable}
		\begin{tabular}{lcccccc}
			\hline
			\multicolumn{1}{l}{\multirow{2}{*}{Method}} & \multicolumn{1}{c}{\multirow{2}{*}{mAP}} & \multicolumn{5}{c}{Category}                                                        \\ \cline{3-7}
			\multicolumn{1}{l}{}
			& \multicolumn{1}{c}{}
			& \multicolumn{1}{c}{FO} & \multicolumn{1}{c}{ST} & \multicolumn{1}{c}{SC} & \multicolumn{1}{c}{UT} & \multicolumn{1}{c}{MU} \\ \hline
			SSD \cite{liu2016ssd} & 70.89  & 76.91  & 35.02  & 93.41    & 65.87& 83.27 \\
			\hline
			+Concate  &72.32 & 79.00  & 36.46  & 94.13  & 68.85    & 83.18 \\
			+EG  & 72.75 & 80.26  & 35.54 & 94.81  & 67.96  & \textbf{85.19}    \\
			+EG+MA/$\mathcal{G}$ & 73.12 & 79.94 & 38.58 & 93.39    & 69.40   & 84.28    \\
			+EG+MA  & 74.01  & \textbf{81.37} & 41.50  & 95.12  & 68.21  & 83.83 \\
			\textbf{+EG+MA+O}  & \textbf{74.27}  & 81.06 & \textbf{42.45}  & \textbf{95.37}  & \textbf{70.31}  & 82.16 \\
			\hline
		\end{tabular}\label{Ablation-result}
		\begin{tablenotes}
			\footnotesize
			\item $^*$Note that ``Concate'' refers to simply concating the edge image and the corresponding original image, ``MA/$\mathcal{G}$'' refers to the Material Awareness module without the Gated convolutional network and ``+EG+MA+O" refers to the model ``DOAM-O''.
		\end{tablenotes}
	\end{threeparttable}
\end{table}
	
	Tab. \ref{Ablation-result} shows that EG improved the performance by $0.43\%$, compared with the strategy of simply concatenating the input image and the corresponding edge image without any other operations of EG. We speculate that it is mainly because the EG gives specific emphasis to edge information via optimizing the loss function, which contributes to the ability to focus more on prohibited items. While simply concatenating operates all the objects for feature fusion in the image equally, even the object is not what we desire to detect.
	
	Meanwhile, model integrating both EG and MA performs better than integrating EG alone by $0.37\%$, which unequivocally proves the effectiveness of MA. Note that the size of prohibited item is about 10$\times$10 in average, so we pick 10$\times$10 as the region scale for each position of the feature map.
	
	\begin{figure*}[!t]
	\centering
	\includegraphics[width=\linewidth]{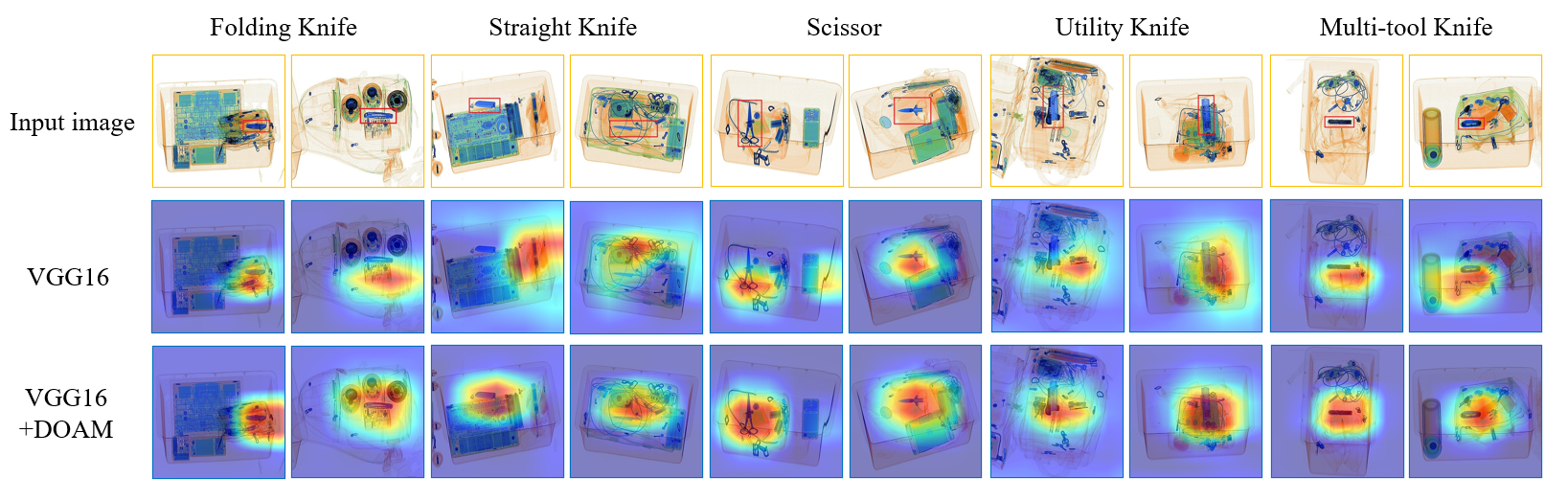}
	\caption{The whole network with Grad-CAM visualization results. We compare the visualization results of the DOAM-integrated network (VGG16+DOAM) with baseline (VGG16). The grad-CAM visualization is calculated for the last convolutional outputs. The ground-truth label is shown on the top of each input image.}
	\label{hotmap}
\end{figure*}

	We choose three different scales of the regions, ($5\times5$, $10\times10$, $15\times15$ respectively), and introduce the gated convolutional neural organization \cite{yu2019free} $\mathcal{G}$ into MA to adaptively select the best feature map, which is generated by average pooling operation with proper pooling size. As a result, the performance improves by $0.9\%$ after drawing $\mathcal{G}$.

	\subsection{Visualization Experiment}

	\begin{figure}[!b]
	\centering
	\includegraphics[width=\linewidth]{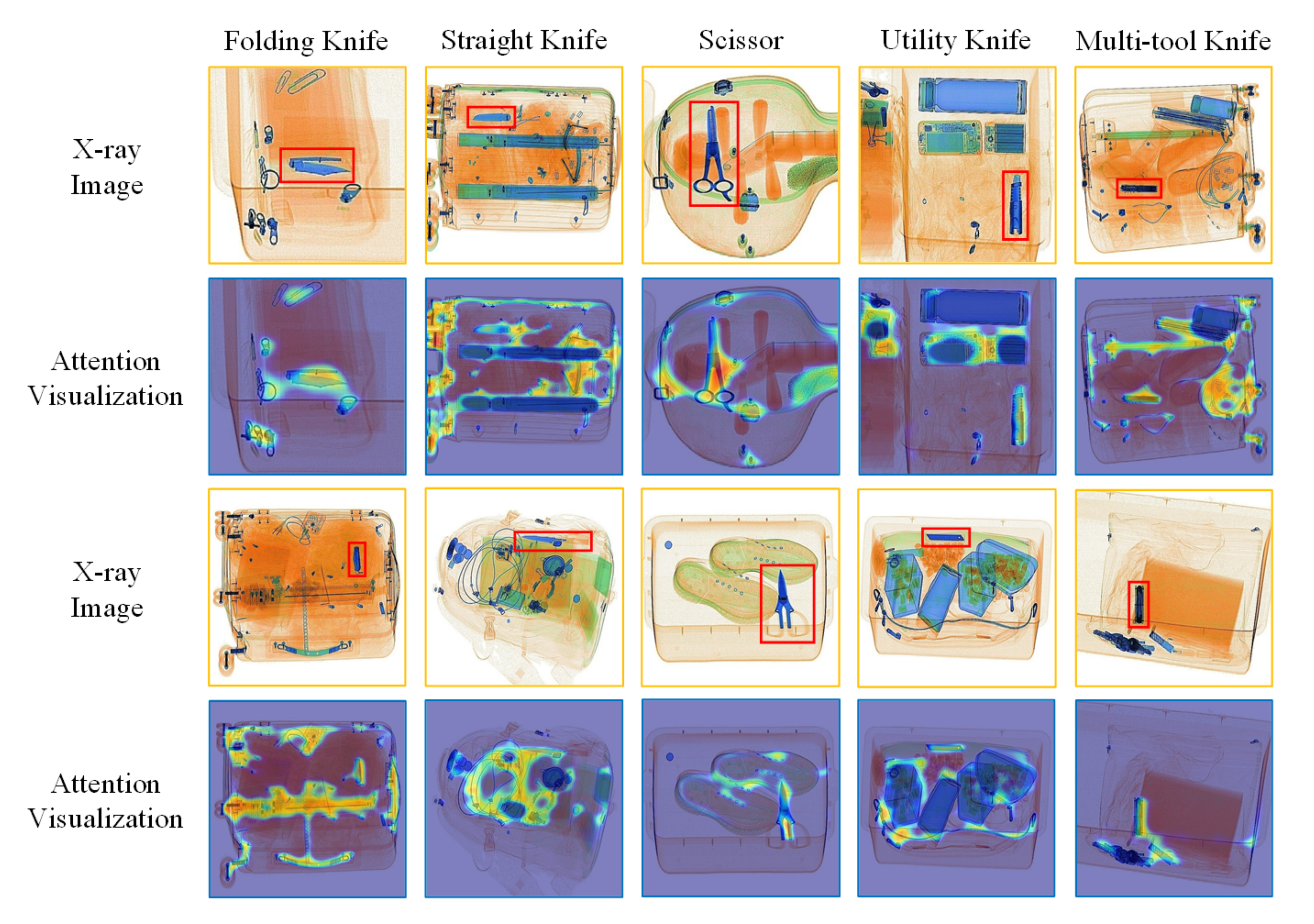}
	\caption{Attention distribution map visualization.}
	\label{att_vis}
\end{figure}

	\subsubsection{Attention Distribution Map Visualization}
	In this section, we visualize the attention map in Fig. \ref{att_vis}, which generated by DOAM to have an intuitive observation of how DOAM works. We showcase 10 input Xray images (two for each class) in rows 1 and 3, as well as the corresponding attention visualizations in rows 2 and 4.
	It is notable that edge and region information are captured by DOAM accurately. For example, in column 4, a red box is marked on a utility knife in the X-ray image (in row 1), and the boundaries of the utility knife are clearly portrayed in the attention visualization (in row 2). Moreover, in the first column, a red box is marked on a folding knife and the corresponding attention map (in row 2) highlights most of the areas where the folding knife lies on. To sum up, visualizations further demonstrate that our proposed model is effective in collecting edge and region information to improve feature representation in occluded prohibited items detection.
	
	\subsubsection{The Whole Network Visualization with Grad-CAM}
	In this section, we apply the Grad-CAM \cite{selvaraju2017grad} to the two networks, VGG16 network \cite{simonyan2014very} and DOAM-integrated network (VGG16+DOAM), and using images from the OPIXray dataset. As we can clearly observe from Fig. \ref{hotmap} that the Grad-CAM masks of DOAM-integrated network cover the target object regions more widely than the single VGG16 network \cite{simonyan2014very}, which verifies the effectiveness of Material Awareness (MA) for utilizing the RIA to aggregate the region information. And also the Grad-CAM masks circle the shape and cover the target more evenly, due to the emphasized edge information by Edge Guidance (EG). From these observations, we verify that the feature refinement process of our model eventually leads the networks to perform better.

	\section{Conclusion}\label{Section:conclusion}
	In this paper, we contribute the first high-quality dataset named OPIXray, which aims at occluded prohibited items detection in security inspection. The images of the dataset are gathered from an airport and these prohibited items are annotated manually by professional inspectors to simulate the real security inspection scene to the greatest extent. Besides, we introduce abundant analysis and discussions for the OPIXray dataset, including construction principles, quality control procedures and potential tasks. To address the occlusion problem in noisy X-ray images, we propose an over-sampling de-occlusion attention network (DOAM-O). To the best of our knowledge, this is the first work that explicitly exploits attention mechanisms and over-sample technologies to prohibited items detection, which might provide a new way to solve the occlusion problem under noisy data scenarios. We hope that our contributions can be beneficial to promote the development of prohibited items detection in noisy X-ray images.
	
	\section{Acknowledgement}
	This work was supported by National Natural Science Foundation of China (62022009, 61872021), Beijing Nova Program of Science and Technology (Z191100001119050), and State Key Lab of Software Development Environment (SKLSDE-2020ZX-06).

	\ifCLASSOPTIONcaptionsoff
	\newpage
	\fi

	
	
	
	\bibliographystyle{IEEEtran}
	\bibliography{ref}
	
	
	% \end{thebibliography}
	

\end{document}